\documentclass[10pt,twocolumn,letterpaper]{article}
\usepackage[pagenumbers]{cvpr}
\usepackage{booktabs}
\usepackage{subcaption}
\usepackage{amsmath,amssymb,bm}
\usepackage{booktabs,multirow}
\usepackage{graphicx}
\usepackage{xcolor}
\usepackage{url}
\usepackage{tikz}
\usetikzlibrary{arrows.meta,positioning,calc,fit}
\usepackage[pagebackref=false,breaklinks,colorlinks,allcolors=cvprblue]{hyperref}
\definecolor{cvprblue}{rgb}{0.21,0.49,0.74}
\definecolor{ember}{RGB}{232,93,42}
\definecolor{tealc}{RGB}{15,139,141}
\definecolor{frz}{RGB}{54,88,160}

\def\paperID{}
\def\confName{IROS}
\def\confYear{2026}

\title{OpenSpace Lab Solution to the IROS 2026 Indoor Exploration Competition}

\author{
Yuxuan Zhang$^{1}$ \qquad 
Dong Li$^{2,3}$ \qquad 
Zezhou Sun$^{4}$ \qquad 
Yuxuan Xu$^{1}$ \\
Siyu Teng$^{5}$ \qquad 
Yuchen Li$^{6}$ \qquad 
Jianjian Yang$^{1}$ \qquad 
Long Chen$^{1,3}$\\
\\
\normalsize $^{1}$China University of Mining and Technology-Beijing\qquad 
$^{2}$Macau University of Science and Technology \\
\normalsize $^{3}$Institute of Automation, Chinese Academy of Sciences \qquad 
$^{4}$Mohamed bin Zayed University of Artificial Intelligence \\
\normalsize $^{5}$Shenzhen University\qquad 
$^{6}$Technical University of Munich
\thanks{Email: doongli@ieee.org}
}

\begin{document}

\maketitle

\begin{abstract}

This report presents the \textbf{OpenSpace Lab}'s solution to the Competition on Intelligent Information Gathering for Single and Multi-Robot Systems Workshops, organized as part of the IEEE/RSJ International Conference on Intelligent Robots and Systems (IROS) 2026. Our team reached 1st place in the Single-Robot Public Track and 3rd place in both the Single- and Multi-Robot Private Tracks.
The single-robot framework utilizes pre-trained map completion predictions for global planning to prioritize unexplored areas. To reconcile map coverage with limited operation time, we introduce a remaining-time-based exploration strategy that integrates homing constraints into the decision-making process.
For multi-robot exploration, we utilize a utility-driven target selection strategy that balances observation gains, movement costs, and budget constraints, leveraging shared map and intent data to eliminate redundant search and maximize coordination efficiency.
Our solution reached a 61.04\% coverage rate in the Single-Robot Public Track, while reaching 39.53\% and 39.91\% coverage in the Single- and Multi-Robot Private Tracks, respectively.
An extended full-length paper based on this report is currently being prepared for submission, and the source code will be released upon acceptance of the full manuscript at \url{https://github.com/OpenSpace-Lab/Indoor-Exploration-IROS2026}.
\end{abstract}

\section{Introduction} \label{sec:intro}

Robotics exploration is a fundamental capability for autonomous robots, aiming to guide robotic actions through the real-time evaluation of information gain to maximize environmental information collection within minimal temporal budgets~\cite{placed2023survey}. This capability underpins a wide spectrum of applications, ranging from indoor search and wild rescue to underwater surveying and planetary exploration~\cite{li2026mining}. Progressively, the target of quantified information extends beyond conventional geometric representations, such as point clouds and signal sources, to high-level spatial semantic representations and field-based environmental resources~\cite{li2026he3,deng2025best}.

Despite extensive research in autonomous robotic exploration, existing methods predominantly treat execution time or path length as a passive evaluation metric after mission completion, rather than an active constraint governing real-time decision-making. In many critical real-world scenarios, such as battery-constrained rovers exploring the lunar south pole or search-and-rescue agents operating in hazardous environments, time-bounded exploration is paramount. Under these strict temporal budgets, robots must not only maximize information gathering within a finite horizon, but also continuously evaluate the time cost of returning to a base station or communications hub to deliver collected data. However, incorporating explicit, hard time constraints into real-time frontier evaluation and path planning remains largely unaddressed in current exploration frameworks~\cite{datta2023prioritized,ericson2025information}.

The Indoor Exploration Competition\footnote{\url{https://github.com/BYU-FROST-Lab/indoor-exploration-competition}}, hosted by Seungchan Kim and Brady Moon at the IROS 2026 Workshop on Intelligent Information Gathering for Single and Multi-Robot Systems\footnote{\url{https://frostlab.byu.edu/IIG-workshop/}}, evaluates autonomous agents under realistic operational constraints. The competition provides scale-dependent step budgets and strictly requires robots to return to the base station range to upload gathered map data.

\begin{figure*}[t]
  \centering
  \includegraphics[width=\linewidth]{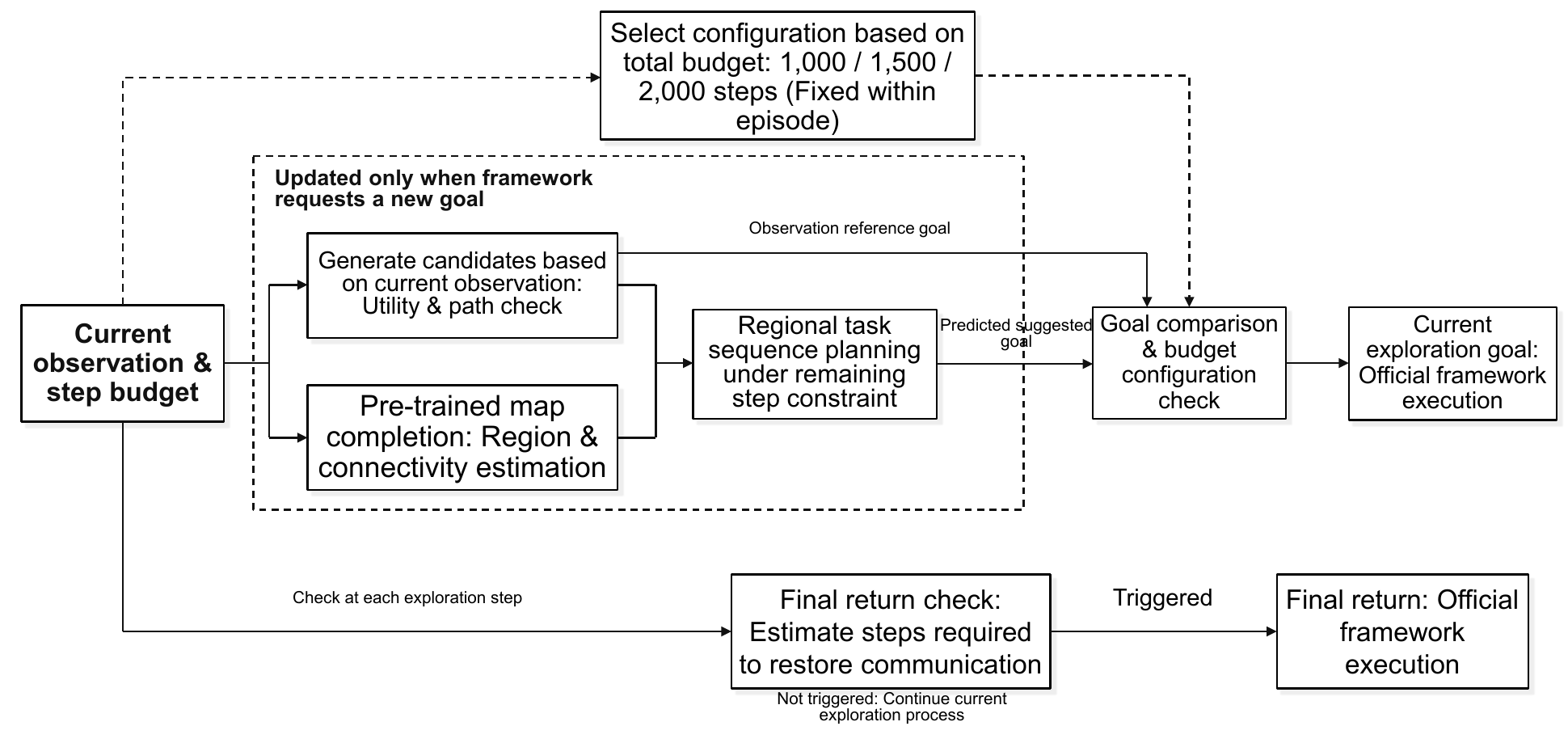}
  \caption{Framework of Single-Robot Exploration.}
  \label{fig:single_robot_framework}
\end{figure*}

To address these challenges, our team, OpenSpace Lab, developed specialized solutions for both single- and multi-robot tracks. The single-robot strategy generates candidate goals based on immediate observations and leverages pre-trained map completion to prioritize exploration sequences. Adapting to varying step budgets, it dynamically adjusts goal selection and pathing rules. The multi-robot extension reduces redundant search by initializing dispersed exploration directions and incorporating shared teammate goals alongside historical trajectories. Both strategies monitor remaining steps against estimated homing costs to guarantee timely return and data delivery. Detailed formulations of these methods are presented in Section~\ref{sec:method}.

\begin{figure*}[t]
  \centering
  \includegraphics[width=\linewidth]{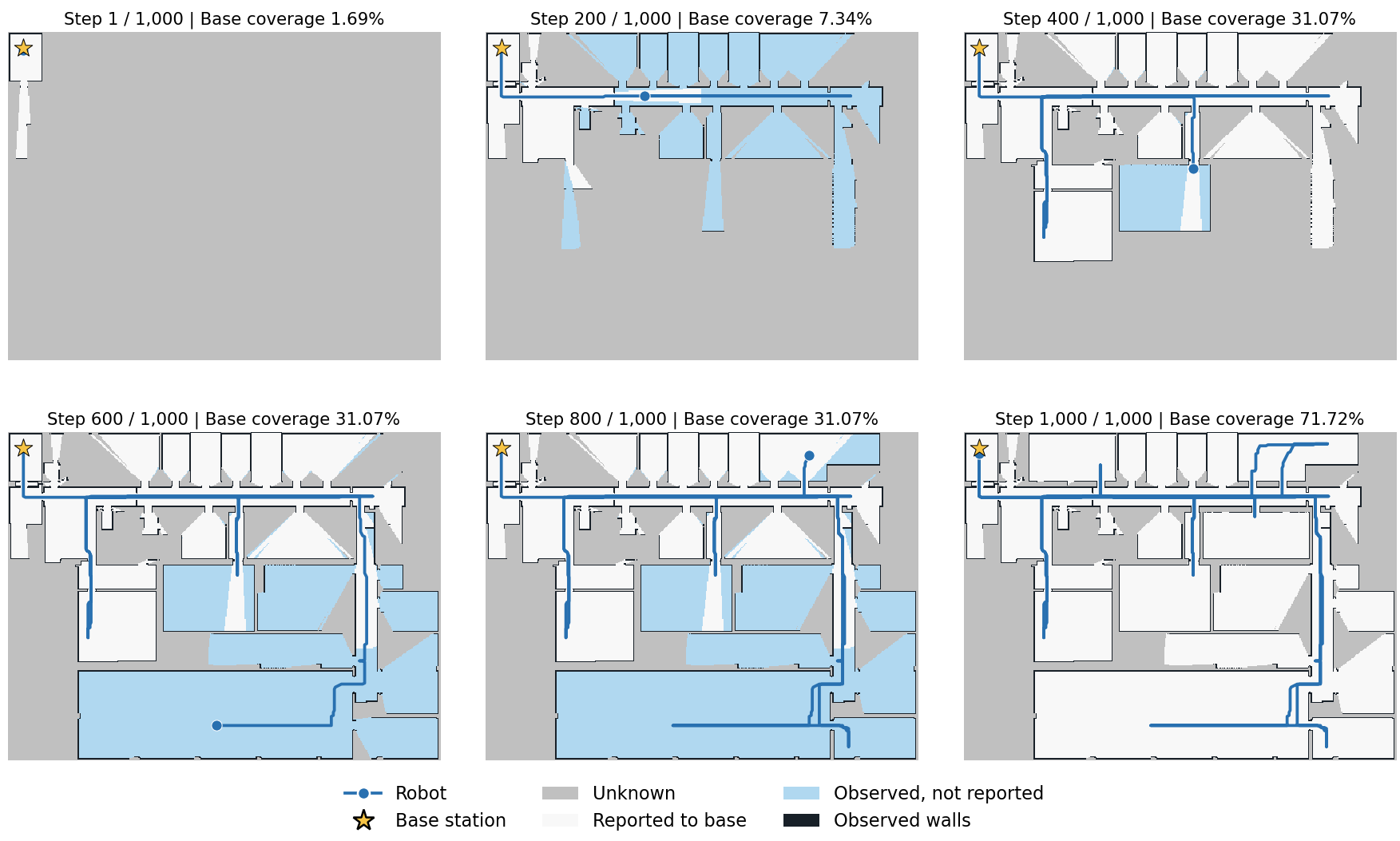}
  \caption{Qualitative visualization of single-robot exploration performance in the first map among the seven public maps.}
  \label{fig:single_robot_exploration}
\end{figure*}

\section{Method} \label{sec:method}

To address the challenges posed by both single- and multi-robot exploration scenarios, the approach developed by OpenSpace Lab adopts a customized two-tier architecture adapted to their distinct operational dynamics. Both pipelines couple core algorithmic formulations with lightweight engineering heuristics to maximize exploration efficiency under strict step budgets.

\subsection{Single-Robot Method} \label{subsec:single_robot}

Our single-robot exploration strategy combines learning-augmented global task ordering with fine-grained local planning, while dynamically adapting execution heuristics across different step budgets. The overall framework is depicted in Figure~\ref{fig:single_robot_framework}.

\begin{figure*}[t]
  \centering
  \includegraphics[width=\linewidth]{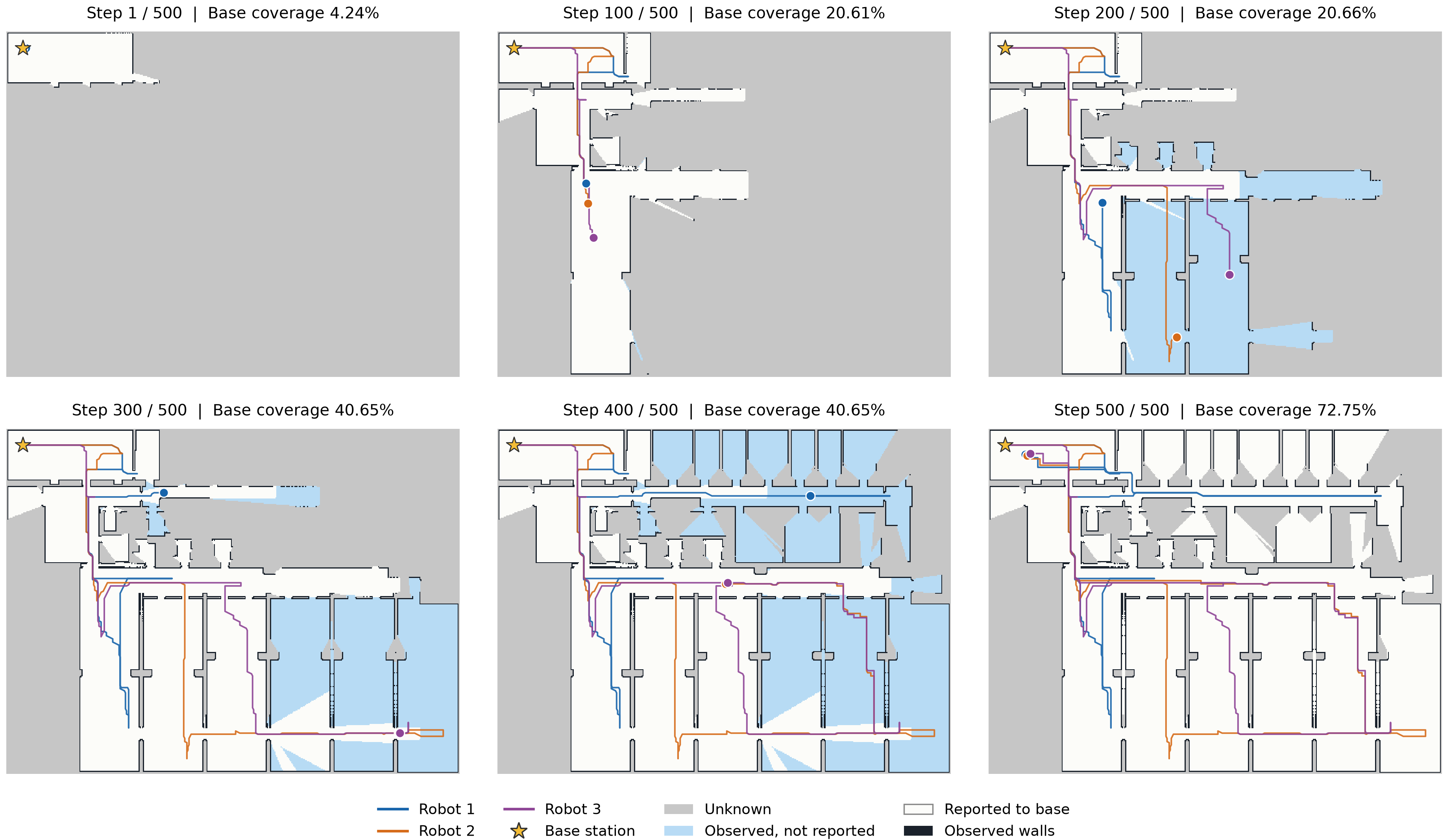}
  \caption{Qualitative visualization of three-robot cooperative exploration in the second map among the seven public maps.}
  \label{fig:multi_robot_exploration}
\end{figure*}

\subsubsection{Global Task Planning and Local Execution}

To guide long-horizon exploration, the global planner utilizes the pre-trained LaMa model~\cite{suvorov2022resolution} within MapEx~\cite{ho2025mapex} for neural map completion. The prediction runs upon initial target selection and updates every eight decision steps, leaving observed free and occupied grids unchanged. Candidate frontier locations are geometrically organized into macro-region tasks (up to seven) labeled by openness, shape, and narrow connectivity. To evaluate visiting sequences $\pi = (q_0, \dots, q_{L-1})$, a beam search (width 36, max length 6) maximizes the sequence score:
\begin{equation}
S(\pi) = \sum_{k=0}^{L-1} 0.92^k U(q_k) - \lambda_{\text{phase}} \sum_{k=1}^{L-1} \mathcal{R}(q_{k-1}, q_k),
\label{eq:seq_score}
\end{equation}
where $U(q_k)$ accounts for information gain, predicted structure, travel cost, and homing burden, while $\mathcal{R}(q_{k-1}, q_k)$ penalizes topological transition risks. Budget feasibility is enforced during beam expansion by verifying cumulative travel, region exploration estimates, and homing overhead against the remaining step budget. Local planning then validates the selected target against an observation-based baseline and issues immediate waypoints, restricting rapid target switching.

\subsubsection{Budget-Aware Engineering Heuristics}
To maximize efficiency under varied step limits (1,000, 1,500, and 2,000 steps), we incorporate adaptive engineering heuristics into target selection and pathing. For short-budget scenarios (1,000 steps), the policy actively identifies alternative goals when route overlap exceeds $75\%$ ($\ge 80$ overlapping steps) to minimize back-tracking overhead. Medium-budget runs (1,500 steps) scale the overlap threshold to $16\%$ ($240$ steps) and defer lateral branch exploration until after the initial $30\%$ phase ($450$ steps), enforcing strict line-of-sight return path checks under $5\times 5$ obstacle inflation. Long-budget settings (2,000 steps) adjust late-stage target sorting by integrating a distance-advantage feature derived from coarse-grid geodesic distances, scaled by a late-stage weight of $0.18$.

\subsubsection{Homing Condition and Data Upload}
Final homing is triggered when the remaining budget falls below the estimated return cost plus a safety margin (10 steps). The homing cost is evaluated as:
\begin{equation}
H = \max\left(H_{\text{path}}, \max\left(0, H_{\text{base}} - c_{\text{core}}\right)\right) + 1,
\label{eq:homing_cost}
\end{equation}
where $H_{\text{path}}$ denotes steps to the first LOS-communicable grid, $H_{\text{base}}$ is the distance to the base station grid, and $c_{\text{core}}$ offsets the distance to the cached reachable communication core. Communication is restored as soon as line-of-sight to the base station is re-established, allowing immediate map upload before physically reaching the base grid.

\begin{table*}[t]
  \centering
  \caption{Results of the single-agent and multi-agent private tracks.}
  \label{tab:private_tracks}
  \begin{subtable}[t]{0.48\linewidth}
    \centering
    \caption{Single-Agent Private Track}
    \label{tab:single_agent_private}
    \begin{tabular}{clc}
      \toprule
      Rank & Team & Score \\
      \midrule
      1 & MobilityAI & 40.26\% \\
      2 & Xplore & 39.65\% \\
      3 & \textbf{OpenSpaceLab} & \textbf{39.53\%} \\
      4 & DZT328 & 37.41\% \\
      5 & HORIZON & 36.63\% \\
      6 & HeRoLab & 35.78\% \\
      7 & Expedition33 & 35.55\% \\
      8 & TAMU-UADY-Robotics & 35.35\% \\
      9 & FAIR-KAIST & 27.70\% \\
      10 & BlueWaspDevTech & 24.17\% \\
      \bottomrule
    \end{tabular}
  \end{subtable}
  \hfill
  \begin{subtable}[t]{0.48\linewidth}
    \centering
    \caption{Multi-Agent Private Track}
    \label{tab:multi_agent_private}
    \begin{tabular}{clc}
      \toprule
      Rank & Team & Score \\
      \midrule
      1 & HeRoLab & 42.64\% \\
      2 & Xplore & 41.12\% \\
      3 & \textbf{OpenSpaceLab} & \textbf{39.91\%} \\
      4 & MobilityAI & 39.24\% \\
      5 & TAMU-UADY-Robotics & 39.05\% \\
      6 & DZT328 & 37.60\% \\
      7 & Expedition33 & 37.38\% \\
      8 & HORIZON & 34.31\% \\
      9 & BlueWaspDevTech & 32.78\% \\
      10 & FAIR-KAIST & 31.27\% \\
      \bottomrule
    \end{tabular}
  \end{subtable}
\end{table*}

\begin{table}[t]
  \centering
  \caption{Results of the single-agent public track.}
  \label{tab:single_agent_public}
  \begin{tabular}{clc}
    \toprule
    Rank & Team & Score \\
    \midrule
    1 & \textbf{OpenSpaceLab} & \textbf{61.04\%} \\
    2 & HeRoLab & 60.12\% \\
    3 & TAMU-UADY-Robotics & 55.89\% \\
    4 & DZT328 & 32.45\% \\
    5 & HORIZON & 25.69\% \\
    \bottomrule
  \end{tabular}
\end{table}

\subsection{Multi-Robot Extension} \label{subsec:multi_robot}

Building upon our single-robot methodology, the multi-robot strategy extends localized exploration to fleet-wide coordination by incorporating shared maps and planned paths, target adjustments based on teammate information, and decentralized dispersion. Local and far-range candidate viewpoints are extracted from obstacle-inflated space and frontier representatives, evaluated via ray-casting information gain $G(q)$ and dual-map $A^*$ path costs $C(q)$. To minimize redundant exploration, candidates are penalized based on their proximity to current and stale teammate intent points $P_{\text{intent}}(q)$ and historical trajectories $P_{\text{traj}}(q)$, while a directional momentum bonus $B(q)$ discourages frequent oscillations. At mission departure, agents preferentially disperse into angular sectors centered at the base station to reduce initial spatial overlap. Furthermore, candidate selection preferentially retains viewpoints satisfying $\tau(q) \le R - H - M$ to reserve steps for homing~\cite{kim2026proid}. If no candidate in a pool satisfies this condition, the candidate with the smallest estimated travel time is retained.

To maintain communication under budget constraints, the fleet combines event-triggered data relays with dynamic handoffs. When the number of unreported cells not delegated to a teammate reaches threshold $C_0$, the minimum interval since the previous ordinary relay trigger has elapsed, the remaining exploration allowance is below a preset threshold, and a connected teammate closer to the base station than the agent is available, an agent initiates an ordinary relay towards the base station. If no such teammate is available, a higher data threshold and a sufficiently low estimated return cost are additionally required. During ordinary relaying, if a currently connected teammate positioned closer to the base station is detected, the agent selects the one nearest to itself, transfers its relay burden via dynamic handoff and immediately resumes frontier exploration. The multi-robot return estimate is $H = \max(H_{\text{comm}}, H_{\text{base}} - 20) + \delta$, where $H_{\text{comm}}$ denotes the estimated number of steps to the first position along the return path where communication with the base station is predicted to be available, and $H_{\text{base}}$ denotes the estimated number of steps to the base station cell. The offset is $\delta=0$ when a path through known free space is available and $\delta=80$ when a path allowing traversal of unknown cells is used instead. When the budget reaches critical limits ($R \le H + M$) or no return path can be found, agents switch to final homing mode, following a planned path towards the base station to restore connectivity and complete map transmission without further relay handoffs.

\section{Experiments} \label{sec:exp}

\noindent\textbf{Evaluation Setup.} 
According to the official evaluation protocol, the preliminary stage used public maps that allowed iterative testing and tuning, whereas the final benchmark utilized unreleased private maps with a single submission attempt. The maps were categorized into small, medium, and large-scale environments, each evaluated across 4 distinct starting locations. In multi-robot runs, all robots and the base station co-located at the same starting position. The single-robot track was assigned step budgets of 1,000, 1,500, and 2,000 for small, medium, and large maps, respectively, while the multi-robot track evaluated teams of 2, 3, 4, and 5 robots with budgets of 500, 1,000, and 1,500 steps, averaging performance across all fleet sizes.

\noindent\textbf{Performance Summary.}
In the preliminary stage across seven public maps, our OpenSpace Lab solution achieved an average coverage rate of 61.04\%, ranking 1st overall (Table~\ref{tab:single_agent_public}) and notably outperforming the 2nd-place team by 3.53 percentage points on large-scale maps (49.81\% vs. 46.28\%). Qualitative exploration results for single- and multi-robot systems on these public maps are illustrated in Figure~\ref{fig:single_robot_exploration} and Figure~\ref{fig:multi_robot_exploration}, respectively. In the final evaluation across ten unreleased private maps, our framework placed 3rd in both tracks (Table~\ref{tab:private_tracks}), while securing 1st place in the single-robot medium-scale map category with a 42.53\% coverage rate.

\section{Conclusion} \label{sec:conclusion}

This report presents the single- and multi-robot exploration strategies developed by OpenSpace Lab, along with our competition results. We are currently preparing a comprehensive algorithm design and full manuscript for formal peer review. Both the competition codebase and the enhanced algorithm implementations will be open-sourced upon paper acceptance.

\section*{Acknowledgement}

We express our sincere gratitude to Seungchan Kim, Brady Moon, and all the organizers for their effort in hosting and maintaining this workshop and competition.

{\small
\bibliographystyle{ieeenat_fullname}
\bibliography{refs}
}
\end{document}